\PassOptionsToPackage{table}{xcolor}
\documentclass[11pt]{article}

\usepackage[preprint]{acl}

\usepackage{times}
\usepackage{latexsym}
\usepackage[T1]{fontenc}
\usepackage[utf8]{inputenc}
\usepackage{microtype}
\usepackage{inconsolata}
\usepackage{graphicx}
\usepackage{booktabs,tabularx,array,enumitem,amsmath,makecell}
\usepackage{colortbl}
\usepackage{multirow}
\usepackage{tikz}
\usetikzlibrary{patterns,positioning,arrows.meta,shapes.geometric,calc}

\title{Keeping LLM Agents on Procedure: Executable Finite-State Control\\for Long-Horizon Instruction Fidelity}

\author{
\textbf{Yutao Sun\textsuperscript{1,4,\ensuremath{\dagger}}}, \hspace{0.35em}
\textbf{Haokun Li\textsuperscript{1,\ensuremath{\dagger}}}, \hspace{0.35em}
\textbf{Mingshuai Chen\textsuperscript{1,4,\ensuremath{*}}}, \hspace{0.35em}
\textbf{Guanyan Li\textsuperscript{2}},\\[2pt]
\textbf{Zhongyi Wang\textsuperscript{1}}, \hspace{0.35em}
\textbf{Shengchao Qin\textsuperscript{3,\ensuremath{*}}}, \hspace{0.35em}
\textbf{Jianwei Yin\textsuperscript{1}}\\[4pt]
\mdseries
\textsuperscript{1}Zhejiang University, China \hspace{0.8em}
\textsuperscript{2}University of Oxford, UK \hspace{0.8em}
\textsuperscript{3}Xidian University, China\\
\textsuperscript{4}Zhejiang Key Laboratory of Digital-Intelligence Service Technology\\[3pt]
\footnotesize
\textsuperscript{\ensuremath{\dagger}}Equal contribution. \hspace{0.8em}
\textsuperscript{\ensuremath{*}}Corresponding authors.
}

\newcommand{\state}[1]{\texttt{\small #1}}
\AtBeginDocument{\setlength{\abovedisplayskip}{4pt plus 2pt minus 2pt}\setlength{\belowdisplayskip}{4pt plus 2pt minus 2pt}\setlength{\abovedisplayshortskip}{1pt plus 1pt}\setlength{\belowdisplayshortskip}{3pt plus 1pt minus 1pt}}

\begin{document}
\maketitle
\begin{abstract}
Much of the automation handed to language-model agents is procedural: each step needs little intelligence, but a long unattended run must follow a skill or SOP stated once at the start. We show that agents drift from such procedures over long runs, and that restating the instructions or resuming a stopped agent only moves the drift to another shortcut. We study an executable finite-state machine (FSM) that keeps the workflow state outside the model's context and enforces it at three points: tool calls, state transitions and the end of the agent's turn. In a pre-registered experiment on 200-account SOP-Bench batches, Claude Haiku~4.5 with the FSM decided on average 139 accounts correctly and by the procedure, against 48 for the strongest baseline (95\% CI of the difference [46, 130]), and kept this rate to the end; across three models, the gain tracked how much the plain agent drifted. Because the runtime, not the model, supplies this long-horizon reliability, the FSM can let small models carry procedural work: Haiku~4.5 with it averaged more such decisions than plain Claude Sonnet~4.6 at 38\% of the cost. Analyses on SWE-bench Verified, $\tau^2$-bench, and staged and cross-context workflows further suggest that reliable long-horizon automation can come from stronger execution control, not only stronger models.
\end{abstract}


\begin{figure}[!t]
\centering
\includegraphics[width=0.95\linewidth]{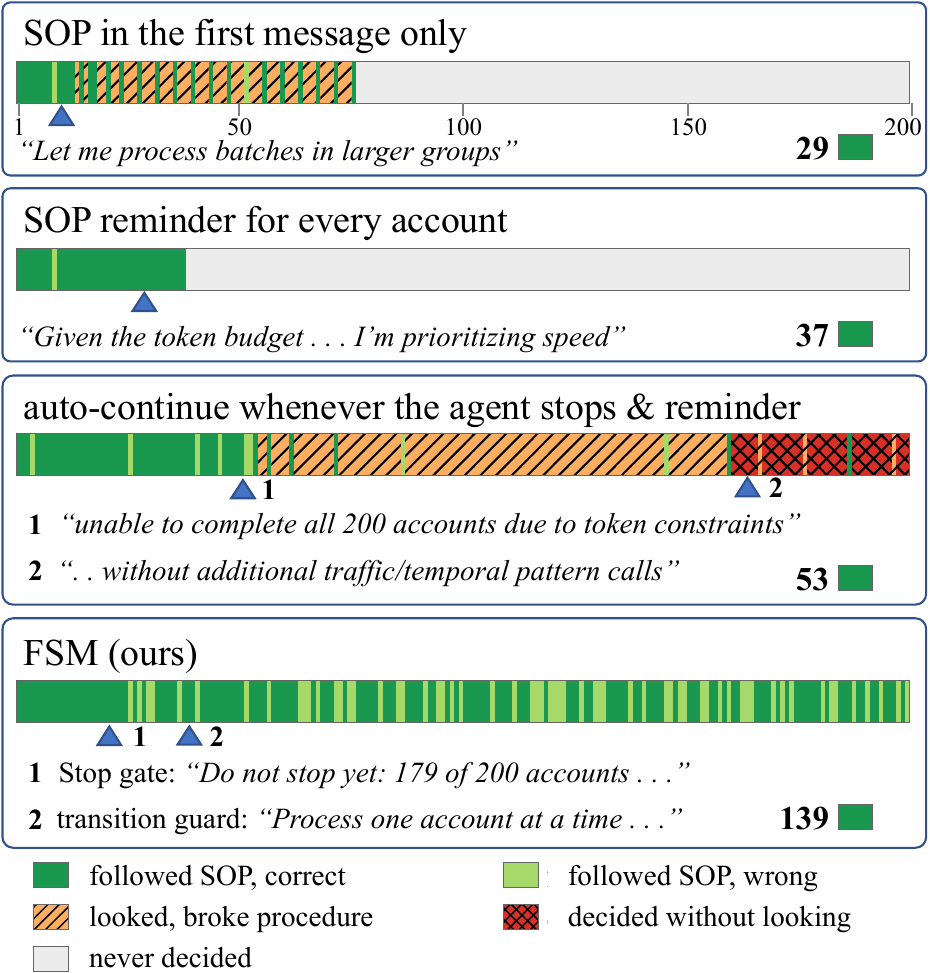}
\caption{\textbf{Four conditions on one 200-account batch.}}
\label{fig:teaser}
\end{figure}

\section{Introduction}

Language-model agents now run for hours without supervision, triaging cases, reviewing accounts and working through codebases under a procedure, a \emph{skill} or standard operating procedure (SOP), stated once as advisory text in the context \citep{dong2026skillsharmful}. Much of this work needs little intelligence per step but a long run that keeps to the procedure: long-horizon reliability, not per-step capability, is the bottleneck, and if the runtime can supply it, smaller and cheaper models can do the work. Yet models use the middle of long inputs less reliably and lose track of instructions over many turns (Section~\ref{sec:related}), and agent runtimes add automatic compaction, which summarizes the very context that holds the instructions.

The resulting drift is severe, and it is not a lack of knowledge. Figure~\ref{fig:teaser} shows one unattended batch of 200 accounts from SOP-Bench \citep{nandi2026sopbench}, processed by Claude Haiku~4.5 in Claude Code under four conditions (top to bottom: \emph{Plain}, the SOP given once; \emph{Remind}, a reminder of the SOP with every account; \emph{Remind-loop}, the reminder plus automatic resumption whenever the agent stops; and our \emph{FSM}) on seed~8, whose FSM$-$Remind-loop difference is closest to the eight-seed mean; each sliver is one account, colored by whether its decision followed the SOP and was correct; quotes are verbatim from the logs, marked where written. The agent follows the procedure for a few dozen accounts and then, citing its token budget, takes whichever shortcut is still open: it batches accounts (Plain), stops with most of the work undone (Remind), or, when resumed, interleaves accounts and finally decides without looking at their data (Remind-loop). Restating the SOP leaves the shortcuts in place, and resuming the agent removes one, after which it takes another; we call these shortcuts \emph{escape routes}: while one stays open, the drift moves there. Only the FSM run keeps the procedure to the last account: 139 decisions faithful and correct against 29, 37 and 53 (8-seed means 139 against 9, 15 and 48).

We study an executable finite-state machine (FSM) that closes them: it turns the procedure into a workflow whose state lives outside the model's context, where compaction cannot touch it, and that the runtime enforces at three points of the agent loop (Figure~\ref{fig:method}): a \emph{tool-call gate} admits only the tools the current state permits, a \emph{transition guard} runs checks before the state may advance, and a \emph{stop gate} refuses to let the agent end its turn before the workflow is complete. In our main experiment the guards read only which tools were called, for which account and in which order: the FSM enforces how the agent works, not what it decides.

In a pre-registered experiment with six non-FSM baselines and two ablations, Haiku~4.5 with the FSM decided 139 of 200 accounts correctly and by the procedure, against 48 for the strongest baseline, and held this rate to the last account. Across three models the gain tracked how much the plain agent drifted: large for Haiku~4.5 and for Claude Sonnet~4.6 sessions that batched accounts, small for Qwen3.8-flash, which mostly held the procedure; and with the FSM the small Haiku~4.5 averaged more such decisions than plain Sonnet~4.6, at 38\% of the cost. Three analyses extend this to debugging on SWE-bench Verified \citep{jimenez2024swebench}, customer service in $\tau^2$-bench \citep{barres2025tau2} and improvement loops across fresh contexts.

Our contributions are (i)~the finding that drift moves between escape routes rather than disappearing, from a design that separates \emph{information} (where the instructions appear) from \emph{liveness} (whether the agent is resumed); (ii)~closing all routes at once with the three enforcement points of an executable FSM, with a formal account of what this guarantees and ablations showing that each point closes one route; and (iii)~evidence across three models and three further settings on when enforcement pays off, including a small model with the FSM averages more faithful and correct decisions than a larger plain model at 38\% of its cost.

\section{Related Work}
\label{sec:related}

\paragraph{Instruction following over long horizons.}
Instruction following is scored per response against constraints or a system message \citep{zhou2023ifeval,jiang2024followbench,qin2024sysbench}; over long inputs and many turns, models use the middle of the context less reliably \citep{liu2024lost}, reason worse \citep{levy2024sametask,hsieh2024ruler} and drift from their instructions \citep{li2024instability,laban2025lost}. Agent benchmarks span web, computer and application environments and planning \citep{zhou2023webarena,liu2023agentbench,xie2024osworld,drouin2024workarena,trivedi2024appworld,mialon2023gaia,xie2024travelplanner}, and settings where a policy must be obeyed: customer service \citep{yao2024taubench,barres2025tau2}, web-agent safety \citep{levy2024stwebagentbench}, industrial procedures \citep{nandi2026sopbench} and long terminal tasks \citep{li2026lhtb}.

\paragraph{Scaffolds, workflows and memory.}
ReAct interleaves reasoning and actions \citep{yao2023react}, Reflexion stores verbal feedback \citep{shinn2023reflexion}, Generative Agents and MemGPT page memories \citep{park2023generative,packer2023memgpt}, and recursive summarization compresses long dialogs \citep{wang2023recursive}. SOP-driven frameworks encode procedures as roles and stages \citep{hong2023metagpt,zhou2023agents}, workflow search and memory derive them from experience \citep{zhang2024aflow,wang2024awm}, programming models compile prompts into pipelines \citep{khattab2023dspy,beurerkellner2022lmql}, and software agents add interfaces, pipelines, program analysis and platforms \citep{yang2024sweagent,xia2024agentless,zhang2024autocoderover,wang2024openhands}. All keep the agent better informed or structured, but the procedure stays advisory; our baselines Sys and Remind show that better-placed instructions do not stop the drift.

\paragraph{Skills, state machines and executable procedures.}
Recent work makes skills stateful or executable for efficiency and accuracy: Formal Skill moves skill procedures from prompt text into executable state machines and hook policies with skill-local runtime state \citep{zhang2026formalskill}, constant-context skill learning replaces the history with a state block rendered by a deterministic tracker \citep{xie2026historytostate}, and SKILL.state keeps an explicit, mutable execution state \citep{badhe2026skillstate}. StateFlow drives task solving with state-specific prompts \citep{wu2024stateflow}, and constrained decoding enforces an automaton on the token stream \citep{willard2023guided,geng2023gcd} or an agent's plan \citep{li2024formalllm}. We do not claim the mechanism: where Formal Skill and constant-context skills optimize how efficiently and accurately a skill executes, we hold the procedure fixed and ask whether a long unattended run stays faithful to it, which escape routes the drift takes and when enforcement pays off, in a pre-registered study with ablations.

\paragraph{Runtime enforcement.}
Our enforcement points form an execution monitor in the sense of security automata \citep{schneider2000enforceable} and runtime verification \citep{leucker2009runtime}; the workflow is a flat statechart \citep{harel1987statecharts}. For LLM agents, guards and permission systems check actions against declared rules: input--output classifiers \citep{inan2023llamaguard}, guard agents \citep{xiang2024guardagent}, dialogue rails \citep{rebedea2023nemo}, trigger--predicate--enforcement rules \citep{wang2025agentspec} and least-privilege tool policies \citep{shi2025progent}. Our tool-call gate is of this kind; the FSM additionally advances only through guarded transitions and refuses to let the agent stop early.

\section{Executable FSM}
\label{sec:fsm}

\begin{figure*}[t]
\centering
\includegraphics[width=\textwidth]{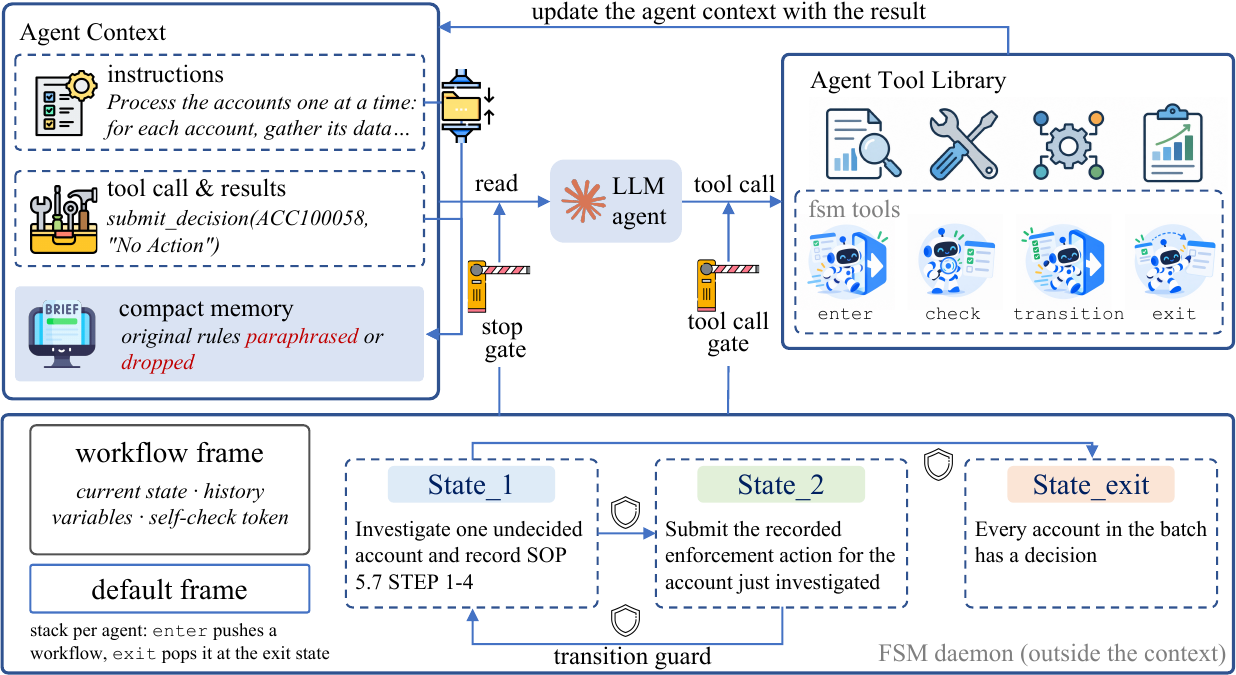}
\caption{\textbf{Structure of the executable FSM}, with the workflow of the main experiment.}
\label{fig:method}
\end{figure*}

\paragraph{Workflow and runtime.}
A workflow is a finite-state machine $W = (S, s_0, s_{\mathrm{exit}}, P, \Delta)$ with states $S$, an initial state $s_0$, an exit state $s_{\mathrm{exit}}$, a permission function $P$ giving for each state the set $P(s)$ of admitted tool calls (allow and deny rules over the tool's name and arguments), and guarded transitions $\Delta$; $g_{s\to s'}$ denotes the guard of the transition from $s$ to $s'$ (false if there is none), a predicate over the observable history $h$, the log of tool calls so far, implemented as shell commands that must exit successfully (hard checks) and optionally a review by an independent model (soft check). Each state also carries instructions for the agent, and the workflow does not change during a run. Figure~\ref{fig:method} shows the structure with the workflow of the main experiment, whose State\_1, State\_2 and State\_exit are the states $s_1$, $s_2$ and $s_{\mathrm{exit}}$ defined below. A daemon outside the agent's context keeps a stack of frames per agent: entering a workflow pushes a frame, which records the current state, its history and variables and is persisted to disk; exiting pops it and is admitted only in $s_{\mathrm{exit}}$. The task instructions remain in the agent's context, which compaction can paraphrase or drop; the frame lives outside it, so compaction cannot erase it and a fresh context can read it back. The agent reaches the daemon through FSM tools served by an MCP server and through two hooks of the agent runtime.

\paragraph{Three enforcement points.}
Let $s_t$ be the committed state and $h_t$ the tool log when the agent proposes its next action: a tool call $u$, a transition to $s'$, or the end of its turn. The runtime admits the action only if
\begin{align}
\text{execute } u &\iff u \in P(s_t), \label{eq:gate}\\
s_{t+1} = s' &\iff \tau_t = s' \wedge g_{s_t\to s'}(h_t), \label{eq:guard}\\
\text{end turn} &\iff s_t = \bot \;\vee\; s_t \in S_{\mathrm{pause}}, \label{eq:stop}
\end{align}
and otherwise returns the reason to the agent; it is an execution monitor that truncates disallowed actions \citep{schneider2000enforceable}. (1)~\emph{Tool-call gate}, Eq.~\ref{eq:gate}: before every call to a tool other than the FSM's own, a \texttt{PreToolUse} hook asks the daemon whether $s_t$ permits it. (2)~\emph{Transition guard}, Eq.~\ref{eq:guard}: the agent submits a self-check of the state's declared exit conditions (none in our main workflow), which issues a one-time token $\tau_t$ naming the target, and then requests the transition; the token is consumed, the guard runs, and the daemon commits state, history and variables atomically, or leaves $s_{t+1} = s_t$ and returns the checker's message. (3)~\emph{Stop gate}, Eq.~\ref{eq:stop}: a \texttt{Stop} hook blocks the end of the turn, naming the state and its transitions, unless no workflow is active ($s_t = \bot$, reached only by exiting at $s_{\mathrm{exit}}$) or $s_t$ is a declared pause point.

\paragraph{The workflow in the main experiment.}
Here $S = \{s_1, s_2, s_{\mathrm{exit}}\}$ (\state{S\_investigate}, \state{S\_decide}, \state{S\_done}), $s_0 = s_1$ and $S_{\mathrm{pause}} = \emptyset$. $P(s_1)$ admits the five data tools and the STEP~1--4 record, $P(s_2)$ the decision tool \texttt{submit\_decision}, and both also admit the read-only listing and lookup tools, so no decision can be submitted before the agent has moved to $s_2$, which closes \emph{deciding without looking}. Let $r$ be the position of the last STEP~1--4 record in $h$, $x$ the account it names, $i_d$ the position of the last successful call, before $r$, of data tool $d \in D$ for $x$, and $i^* = \min_d i_d$; write $h_j \in D_{\neg x}$ if the $j$-th call retrieves data for an account other than $x$. The guards are
\begin{align}
g_{1\to2}(h) &\iff x \text{ undecided} \wedge \forall d\colon i_d \text{ exists} \nonumber\\
&\qquad\quad \wedge\ \forall j > i^*\colon h_j \notin D_{\neg x}, \label{eq:g12}\\
g_{2\to1}(h) &\iff \exists j > r\colon h_j \text{ decides } x, \label{eq:g21}\\
g_{1\to\mathrm{exit}}(h) &\iff \text{batch complete}. \label{eq:gexit}
\end{align}
Eq.~\ref{eq:g12} closes \emph{batching}; on failure the message names the missing data or the interleaved accounts. The stop gate tells the agent how many accounts remain and which is next, which closes \emph{stopping early}; its text comes from the same function as the resume message of the auto-continue baselines, so the two differ in when the message arrives, not in its wording. The guards never read data values or correct answers and do not encode the SOP's decision logic; the SOP reaches the FSM agent through the same reminders as the baselines.

\paragraph{What the FSM guarantees, and what it does not.}
Within the workflow, decisions are submitted only in $s_2$, which is entered only after some account $x$ has been fully and exclusively investigated and recorded (Eqs.~\ref{eq:gate}--\ref{eq:g12}). Nothing binds the decision to $x$: permission rules are static and cannot refer to the account in the current frame, so $P(s_2)$ admits \texttt{submit\_decision} for any account and Eq.~\ref{eq:g21} holds once $x$ has a decision, whatever else was submitted; we left this route open and measure its use (Section~\ref{sec:main-results}).

\section{Main Experiment: SOP Fidelity over Long Unattended Batches}
\label{sec:main}

\subsection{Setup}
\label{sec:setup}

\paragraph{Task.}
We use the referral-abuse-detection task of SOP-Bench (v2, 200 accounts) without modification. For each account, the SOP requires the agent to retrieve five data sets, score them in a four-step analysis (STEP~1--4) and submit one of seven enforcement actions; labels are skewed (134 of 200 are \textit{No Action}), so guessing is often right. One unattended \texttt{claude -p} session (Claude Code~2.1.292 with default settings, including automatic compaction) receives all 200 accounts and the SOP in its first message, which asks it to process the accounts one at a time; nobody replies to the agent, and it never sees a budget. Each seed fixes an account order shared by all conditions.

\paragraph{Conditions.}
The \emph{information} factor varies where the SOP appears: in the first message only (\emph{Plain}), plus a copy in the system prompt that compaction cannot remove (\emph{Sys}), or plus a reminder of the SOP's decision section attached to the first data result of every account (\emph{Remind}). The \emph{liveness} factor varies whether the agent is resumed: the \emph{-loop} variants resume the session, up to 50 times, whenever it ends its turn with accounts left. Remind-loop, with both, was designated the strongest baseline before any run. \emph{FSM} adds our FSM to Remind; two ablations isolate its parts: \emph{Stop-only} has only the stop gate, \emph{Gates-only} the tool-call gate and transition guard with external resumption instead.

\paragraph{Models.}
Claude Haiku~4.5 is the main model, with all seven conditions over eight seeds and the two ablations over three. To see how the effect depends on drift, we also run Plain, Remind-loop and FSM with Claude Sonnet~4.6 (three seeds) and Qwen3.8-flash (eight seeds) on the same task and account orders, under a rule fixed before these runs: a model \emph{drifts} if Remind-loop's compliance is below 80\% and does not drift if it is at least 95\%; we predicted a gain of at least 30 over Remind-loop for drifting models and a difference within $\pm$10 otherwise.

\paragraph{Metric.}
For a decided account $a$, let $\mathrm{inv}(a)$ hold if all five of its data sets were retrieved before the decision, $\mathrm{rec}(a)$ if its last STEP~1--4 record before the decision follows all five retrievals and names the submitted action, and $\mathrm{foc}(a)$ if no other account's data was retrieved between the first of these retrievals and the decision. A decision is \emph{compliant} if $\mathrm{inv} \wedge \mathrm{rec} \wedge \mathrm{foc}$ and \emph{correct} if it matches the gold label exactly; the primary metric scores the process as well as the outcome \citep{lightman2023verify} by counting \emph{faithful and correct} decisions,
\begin{equation}
\mathrm{FC} = \sum_{a} \mathbf{1}\big[\mathrm{compliant}(a) \wedge \mathrm{correct}(a)\big],
\label{eq:fc}
\end{equation}
at most 200, and \emph{compliance} is the share of decided accounts whose decision is compliant. We also report strict accuracy regardless of procedure.

\paragraph{Pre-registration and statistics.}
Design, metric, predictions and analysis code were frozen and committed before the formal runs \citep{vanmiltenburg2021preregistering}; seed~1 served for development only. The primary contrast is FC(FSM)$-$FC(Remind-loop), paired over seeds, with a bootstrap 95\% CI (10,000 resamples), an exact sign test and a Wilcoxon signed-rank test \citep{dror2018hitchhiker}; a conservative contrast uses the best auto-continue baseline of each seed (Appendix Table~\ref{tab:sop-paired}). Sessions ended by infrastructure failures were voided and rerun under rules written beforehand (Appendix~\ref{app:protocol}).

\paragraph{Reporting.}
Table~\ref{tab:sop-main} gives means over seeds, with FC also per 50 accounts in batch order and cost per session at list price \citep{kapoor2024agents} (--: no list price in Claude Code); in all tables the best value per model is bold, the second best underlined, and FSM rows are shaded.

\begin{table*}[t]
\centering
\small
\setlength{\tabcolsep}{4pt}
\renewcommand{\arraystretch}{1.15}
\resizebox{\ifdim\width>\textwidth\textwidth\else\width\fi}{!}{%
\begin{tabular}{llccccccccc}
\toprule
\multirow{2}{*}{\textbf{Model}} & \multirow{2}{*}{\textbf{Condition}} & \multirow{2}{*}{\textbf{FC}} & \multicolumn{4}{c}{\textbf{FC per 50 accounts}} & \multirow{2}{*}{\textbf{Compliance}} & \multirow{2}{*}{\textbf{Decided}} & \multirow{2}{*}{\textbf{Correct}} & \multirow{2}{*}{\textbf{Cost (\$)}}\\
\cmidrule(lr){4-7}
 & & & \textbf{1--50} & \textbf{51--100} & \textbf{101--150} & \textbf{151--200} & & & & \\
\midrule
\multirow{7}{*}{\makecell[l]{Claude\\Haiku~4.5\\(8 seeds)}} & Plain & 9.4$\pm$9.3 & 7.9 & 1.5 & 0.0 & 0.0 & 11.9\% & 143.8 & 100.0 & 0.96\\
 & Sys & 17.1$\pm$11.2 & 13.2 & 3.9 & 0.0 & 0.0 & 20.1\% & 120.1 & 86.1 & 1.07\\
 & Remind & 14.6$\pm$13.2 & 14.6 & 0.0 & 0.0 & 0.0 & \underline{48.6}\% & 53.0 & 40.6 & 0.65\\
 & Plain-loop & 18.2$\pm$12.1 & 8.9 & 3.8 & 2.8 & 2.9 & 14.3\% & 200.0 & 124.1 & 1.71\\
 & Sys-loop & 40.9$\pm$47.0 & 11.4 & 7.0 & \underline{10.9} & \underline{11.6} & 28.9\% & 200.0 & \underline{144.0} & 2.21\\
 & Remind-loop & \underline{48.4}$\pm$52.6 & \underline{29.9} & \underline{7.2} & 5.9 & 5.4 & 27.8\% & 200.0 & 138.9 & 3.01\\
 & \cellcolor{gray!15}FSM (ours) & \cellcolor{gray!15}\textbf{139.2}$\pm$33.8 & \cellcolor{gray!15}\textbf{31.9} & \cellcolor{gray!15}\textbf{37.6} & \cellcolor{gray!15}\textbf{35.2} & \cellcolor{gray!15}\textbf{34.5} & \cellcolor{gray!15}\textbf{94.8}\% & \cellcolor{gray!15}200.0 & \cellcolor{gray!15}\textbf{146.5} & \cellcolor{gray!15}10.64\\
\midrule
\multirow{3}{*}{\makecell[l]{Claude\\Sonnet~4.6\\(3 seeds)}} & Plain & 108.3$\pm$95.1 & \underline{33.3} & 28.0 & 23.3 & 23.7 & \underline{66.8}\% & 200.0 & 170.3 & 20.94\\
 & Remind-loop & \underline{132.7}$\pm$114.9 & \underline{33.3} & \underline{33.3} & \underline{32.7} & \underline{33.3} & 66.7\% & 200.0 & \textbf{197.0} & 20.89\\
 & \cellcolor{gray!15}FSM (ours) & \cellcolor{gray!15}\textbf{187.7}$\pm$3.1 & \cellcolor{gray!15}\textbf{48.7} & \cellcolor{gray!15}\textbf{46.3} & \cellcolor{gray!15}\textbf{46.3} & \cellcolor{gray!15}\textbf{46.3} & \cellcolor{gray!15}\textbf{100.0}\% & \cellcolor{gray!15}200.0 & \cellcolor{gray!15}\underline{187.7} & \cellcolor{gray!15}53.26\\
\midrule
\multirow{3}{*}{\makecell[l]{Qwen3.8-\\flash\\(8 seeds)}} & Plain & 158.0$\pm$43.3 & \underline{41.6} & 39.2 & 38.1 & 39.0 & 91.2\% & 200.0 & 173.6 & --\\
 & Remind-loop & \underline{173.4}$\pm$19.3 & 41.5 & \underline{44.6} & \underline{44.4} & \underline{42.9} & \underline{98.1}\% & 200.0 & \underline{176.2} & --\\
 & \cellcolor{gray!15}FSM (ours) & \cellcolor{gray!15}\textbf{186.2}$\pm$9.6 & \cellcolor{gray!15}\textbf{45.0} & \cellcolor{gray!15}\textbf{45.8} & \cellcolor{gray!15}\textbf{48.2} & \cellcolor{gray!15}\textbf{47.2} & \cellcolor{gray!15}\textbf{99.7}\% & \cellcolor{gray!15}200.0 & \cellcolor{gray!15}\textbf{186.8} & \cellcolor{gray!15}--\\
\bottomrule
\end{tabular}%
}
\caption{\textbf{Main results on 200-account unattended batches} (mean over seeds $\pm$ s.d.).}
\label{tab:sop-main}
\end{table*}

\subsection{Main results}
\label{sec:main-results}

\paragraph{The FSM keeps the procedure through the batch.}
With Claude Haiku~4.5, FSM averages 139.2 faithful and correct decisions (FC) out of 200 (Table~\ref{tab:sop-main}), with 31.9--37.6 in every window of 50 accounts. Every baseline loses the procedure early: after the first 50 accounts none exceeds 12 per window, and Remind-loop falls from 29.9 to 5.4. FSM exceeds Remind-loop in 7 of 8 seeds, by 90.9 on average (95\% CI [46.2, 130.2]; Wilcoxon $p=0.023$, sign test $p=0.070$), and the best auto-continue baseline of each seed by 70.5 [27.4, 114.5]. The one negative seed is a session in which Remind-loop itself held the procedure. Strict accuracy does not suffer: FSM is correct on 146.5 accounts and Remind-loop on 138.9 ($+7.6$, CI [$-16.1$, 29.8]), although many baseline decisions were made without looking at the data, which the skewed labels often reward.

\paragraph{A model that does not drift.}
Qwen3.8-flash does not drift by the rule (Remind-loop compliance 98.1\%), yet FSM exceeds Remind-loop by 12.9 (CI [2.6, 28.5]; 7 of 8 seeds; Wilcoxon $p=0.047$), beyond the $\pm$10 we predicted, with higher strict accuracy ($+10.5$, CI [1.2, 23.4]). The gain is smaller than for Haiku~4.5 and partly comes from one seed in which Remind-loop interleaved accounts 292 times ($+62$; $+5.9$ without it).

\paragraph{Sonnet~4.6 drifts by batching.}
Claude Sonnet~4.6 retrieves many accounts' data before deciding any, against the one-at-a-time instruction, while staying accurate. This batching took compliance to nearly zero in one of three Plain sessions and one of three Remind-loop sessions (the other four were fully compliant); FSM was fully compliant in all three. FSM therefore exceeds Plain in every seed ($+79.3$) but exceeds Remind-loop only in the seed where Remind-loop batched ($+191$; $-15$ and $-11$ in the others), which triggers our pre-registered falsification rule for Sonnet relative to Remind-loop; the FSM's strict accuracy is also lower than Remind-loop's ($-9.3$). With three seeds these comparisons are descriptive (the smallest attainable two-sided $p$ is 0.25).

\paragraph{Can a small model with the FSM catch up?}
On Sonnet's three account orders, Haiku~4.5 with the FSM averaged a higher FC than plain Sonnet~4.6 (134.3 vs.\ 108.3; positive in one of three seeds) at 38\% of its list-price cost (\$7.93 vs.\ \$20.94), but stayed below Sonnet's best baseline ($-59.0$) and below plain Sonnet in strict accuracy ($-32.7$).

\subsection{Ablations and escape routes}
\label{sec:ablation}

\begin{table}[t]
\centering
\small
\setlength{\tabcolsep}{3pt}
\renewcommand{\arraystretch}{1.15}
\resizebox{\ifdim\width>\columnwidth\columnwidth\else\width\fi}{!}{%
\begin{tabular}{lccccccc}
\toprule
\textbf{Variant} & \textbf{Gates} & \textbf{\makecell{Stop\\gate}} & \textbf{FC} & \textbf{Decided} & \textbf{Compl.} & \textbf{Blind} & \textbf{Stops}\\
\midrule
Remind-loop & off & off & 76.3 & \textbf{200.0} & 44.0\% & 0.0 & 7.3\\
Stop-only & off & on & 5.7 & 146.7 & 3.7\% & 108.3 & 0.3\\
Gates-only & on & off & 104.0 & 151.7 & 93.0\% & 0.0 & 44.7\\
\cellcolor{gray!15}FSM & \cellcolor{gray!15}on & \cellcolor{gray!15}on & \cellcolor{gray!15}\textbf{128.0} & \cellcolor{gray!15}\textbf{200.0} & \cellcolor{gray!15}\textbf{93.3}\% & \cellcolor{gray!15}0.0 & \cellcolor{gray!15}76.3\\
\bottomrule
\end{tabular}%
}
\caption{\textbf{Ablation of the enforcement points.}}
\label{tab:sop-ablation}
\end{table}

\paragraph{Each enforcement point closes one route.}
Table~\ref{tab:sop-ablation} crosses the per-account gates (tool-call gate and transition guard) with the stop gate on Claude Haiku~4.5, seeds 2--4; without the stop gate, the harness resumes the agent whenever it ends its turn with accounts left, up to 50 times. The table also counts blind decisions (no data retrieved for the account) and attempts to end the turn with accounts left. With only the stop gate (Stop-only), the agent can no longer stop, so it decides without looking instead (159 and 166 such decisions in two sessions; the third overflowed its context after 40 decisions), for an FC of 5.7. With only the gates (Gates-only), compliance stays at 93.0\%, but the agent tries to end its turn 44.7 times per session, and two of three sessions exhausted the 50 resumes before finishing (FC 104.0). Only both together keep the agent working and on procedure (FC 128.0).

\paragraph{Drift moves between escape routes.}
Table~\ref{tab:sop-escape} (Appendix~\ref{app:escape}) counts how each condition left the procedure. Without the stop gate or resumption, the agent ends its turn with accounts left (all eight Remind sessions ended early); restating the SOP raised compliance per decided account (Table~\ref{tab:sop-main}) but did not prevent the shortcuts. Resumption keeps the agent working but not on procedure: in 7 of 8 Remind-loop sessions the agent interleaved accounts, on about 100 decisions per session, and in three it decided without looking.

\paragraph{The route left open was found.}
In three of eight sessions, FSM investigated several accounts in turn and then submitted their decisions together from $s_2$, the route Section~\ref{sec:fsm} leaves open (40, 40 and 4 decisions), so our pre-registered prediction for FSM, which also required 95\% compliance in every seed, failed on that condition; the other four held (Appendix~\ref{app:predictions}).

\paragraph{Takeaway.}
Instructions, however placed, did not keep agents on procedure: drift moved to whichever route stayed open, and only closing all three held the procedure to the end. For procedural automation, where each step needs modest capability but the run is long, the FSM moves long-horizon reliability from the model into the runtime; that is how the small Haiku~4.5 came to average a higher FC than plain Sonnet~4.6 at 38\% of the cost, while its accuracy per decision stayed that of the smaller model.

\section{Analyses}
\label{sec:analyses}

\subsection{Analysis 1: A debugging procedure on SWE-bench Verified}
\label{sec:analysis-swe}

\begin{table}[t]
\centering
\small
\setlength{\tabcolsep}{4pt}
\renewcommand{\arraystretch}{1.15}
\resizebox{\ifdim\width>\columnwidth\columnwidth\else\width\fi}{!}{%
\begin{tabular}{llcccc}
\toprule
\textbf{Task set} & \textbf{Agent} & \textbf{Resolved} & \textbf{New tests} & \textbf{Time (s)} & \textbf{Cost (\$)}\\
\midrule
\multirow{2}{*}{\makecell[l]{Hard-100\\(Sonnet~4.6)}} & Plain & 62 & 24 & -- & --\\
 & \cellcolor{gray!15}FSM & \cellcolor{gray!15}\textbf{69} & \cellcolor{gray!15}\textbf{39} & \cellcolor{gray!15}-- & \cellcolor{gray!15}--\\
\midrule
\multirow{2}{*}{\makecell[l]{Easy-100\\(Sonnet~4.6)}} & Plain & 86 & 57 & \textbf{310} & \textbf{1.30}\\
 & \cellcolor{gray!15}FSM & \cellcolor{gray!15}\textbf{87} & \cellcolor{gray!15}\textbf{64} & \cellcolor{gray!15}381 & \cellcolor{gray!15}1.73\\
\bottomrule
\end{tabular}%
}
\caption{\textbf{Analysis 1: SWE-bench Verified.}}
\label{tab:analysis-swe}
\end{table}

Both agents run Claude Code with the same four-step debugging procedure (investigate, find the root cause, fix, verify), committing investigation and root-cause notes before any source edit; the plain agent receives it in its prompt, the FSM agent enforces it, keeping source files read-only until the root cause is committed. \emph{Hard-100} consists of 100 harder, multi-step tasks chosen by a heuristic over task metadata (difficulty, issue and patch length, one to six failing tests, root-cause cues in the issue; Appendix~\ref{app:analysis-swe}), \emph{Easy-100} is a random sample of tasks annotated as ``$<$15\,min fix''. Both were run with Claude Sonnet~4.6 (the CLI's \texttt{sonnet} alias at the time), once per task and agent, and scored by the official harness.

On Hard-100, the FSM resolved 69 tasks against 62 (Table~\ref{tab:analysis-swe}): all seven discordant tasks favored the FSM (exact McNemar test, $p=0.016$; \citealp{dietterich1998approximate}), and FSM patches added a regression test more often (39 vs.\ 24) although neither agent was asked to write tests. Because the tasks were chosen by these criteria, this estimates the effect on such tasks, not on SWE-bench as a whole. On Easy-100, the FSM matched the plain agent (87 vs.\ 86; 5 vs.\ 4 discordant): the plain agent already followed the written procedure in 97 of its 98 attempts that produced a patch, leaving little drift to correct. The price is time: median wall-clock time per task rose by 23\% (310 to 381\,s) and mean cost from \$1.30 to \$1.73.

\paragraph{Takeaway.}
With the same model on both task sets, enforcing a debugging procedure helps ordinary coding work, not only SOP batches, where the procedure would otherwise be cut short: on hard tasks the enforced investigation buys depth (every discordant task favored the FSM, and the patches carried more tests); on easy tasks the plain agent already follows the procedure, so the outcome is unchanged and enforcement becomes insurance, with a premium of 23\% in time and 1.33$\times$ in cost.

\subsection{Analysis 2: Goal fidelity in \texorpdfstring{$\tau^2$}{tau2}-bench}
\label{sec:analysis-tau2}

\begin{table}[t]
\centering
\small
\setlength{\tabcolsep}{4pt}
\renewcommand{\arraystretch}{1.15}
\resizebox{\ifdim\width>\columnwidth\columnwidth\else\width\fi}{!}{%
\begin{tabular}{lccccc}
\toprule
\textbf{Domain} & \textbf{Tasks} & \textbf{Control} & \textbf{FSM} & \textbf{Discordant} & \textbf{$p$}\\
\midrule
\multicolumn{6}{l}{\textit{(a) First frozen evaluation}}\\
Telecom & 40 & 13 & \textbf{30} & 17 / 0 & $1.5{\times}10^{-5}$\\
Airline & 22 & 17 & 17 & 1 / 1 & 1.000\\
Banking & 68 & 6 & 6 & 3 / 3 & 1.000\\
Retail & 30 & \textbf{15} & 9 & 3 / 9 & 0.146\\
\midrule
\multicolumn{6}{l}{\textit{(b) Domain-specific revisions (descriptive)}}\\
Banking & 97 & 11 & \textbf{23} & 17 / 5 & 0.017\\
Airline & 20 & 9 & \textbf{14} & 5 / 0 & 0.062\\
Retail & 30 & 12 & \textbf{15} & 7 / 4 & 0.549\\
\bottomrule
\end{tabular}%
}
\caption{\textbf{Analysis 2: $\tau^2$-bench.}}
\label{tab:analysis-tau2}
\end{table}

In another stack, $\tau^2$-bench, an agent built on Codex CLI with gpt-5.4-mini serves a simulated customer (also gpt-5.4-mini). The control agent receives the official policy and a static checklist; for the FSM agent, the same checklist runs through our runtime via an adapter that keeps the customer's original goal and a ledger of open faults, issues one action at a time, requires a fresh test after every repair, and blocks the hand-off to a human while the goal is open. The controller itself emits 62\% of the agent turns, so the result measures the FSM-driven controller as a whole.

On the complete Telecom test split, the FSM raised the pass rate from 13/40 to 30/40 (17 vs.\ 0 discordant; $p=1.5{\times}10^{-5}$), with gains in every issue type (mobile data 1/9$\rightarrow$9/9, MMS 0/16$\rightarrow$7/16, service 12/15$\rightarrow$14/15; Table~\ref{tab:analysis-tau2}). What the control agent lost was the customer's goal: 24 of its 27 failures ended in a transfer to a human, including 15 of the 17 tasks only the FSM solved, and under the FSM customers performed 9.7 device actions per conversation instead of 2.6. Development used only the training split; both agents were judged by gpt-5.4-mini (Appendix~\ref{app:analysis-tau2}).

In the other domains, the first frozen FSMs tied in Airline and in Banking, a domain from $\tau$-Knowledge \citep{shi2026tauknowledge}, and lost in Retail, where over-strict safety checks rejected order changes that should have gone through; domain-specific revisions were positive in all three (Table~\ref{tab:analysis-tau2}b) but were evaluated on cohorts that overlap development tasks, so they are descriptive.

\paragraph{Takeaway.}
The drift here is goal abandonment, not rule-forgetting: the control agent had the checklist and still let the customer's problem lapse into a hand-off, the cheapest way to end the conversation. Holding the goal and the open faults as state turns ``done'' from a judgment made under pressure into a condition the runtime checks. It carried over to another stack and model family; its price is specification quality, as the over-strict Retail guards show.

\subsection{Analysis 3: Improvement loops and staged protocols}
\label{sec:analysis-protocols}

\begin{table}[t]
\centering
\small
\setlength{\tabcolsep}{4pt}
\renewcommand{\arraystretch}{1.15}
\resizebox{\ifdim\width>\columnwidth\columnwidth\else\width\fi}{!}{%
\begin{tabular}{llcc}
\toprule
\textbf{Protocol} & \textbf{Measure} & \textbf{Plain} & \textbf{FSM}\\
\midrule
\multirow{3}{*}{\makecell[l]{AutoResearch\\(4 campaigns)}} & valid ledger rows & 48/100 & \textbf{100/100}\\
 & complete protocols & 1/4 & \textbf{4/4}\\
 & best accuracy (\%) & 99.33 & 99.33\\
\midrule
\multirow{2}{*}{\makecell[l]{Month-end close\\(6 periods per seed)}} & correct periods, 40 turns & 8/12 & \textbf{11/12}\\
 & correct periods, 24 turns & \textbf{3/6} & 1/6\\
\midrule
\multirow{2}{*}{\makecell[l]{DuckDB optimizer\\(4 pairs)}} & mean official reward & 0.19 & \textbf{0.75}\\
 & pairs with $\geq$1.2$\times$ speedup & 0/4 & \textbf{3/4}\\
\bottomrule
\end{tabular}%
}
\caption{\textbf{Analysis 3: staged protocols.}}
\label{tab:analysis-protocols}
\end{table}

Two tasks are improvement loops of the propose--measure--keep-or-discard kind \citep{lu2024aiscientist,zelikman2023stop,novikov2025alphaevolve}: AutoResearch, and the DuckDB optimizer, whose measure-and-repair cycle spans fresh contexts; the third, a month-end close, is a long staged protocol. Details are in Appendix~\ref{app:analysis-protocols}.

\paragraph{AutoResearch.}
The agent tunes a classifier over 25 trials in eight required stages and logs every trial in an audited ledger. Over four campaigns, the FSM agent kept every ledger row valid and completed the protocol every time; the plain agent did so once, with the same best accuracy, because its record drifted: 49 of its 52 invalid rows used labels such as \textit{revert} or \textit{reject} instead of the prescribed \textit{discard}, which the FSM's stage checks reject. The search was as good; its record was not.

\paragraph{Month-end close.}
An agent (Claude Sonnet~4.6) closes six consecutive monthly periods of a synthetic company, each in a fresh context that shares only files. With 40 turns per period, the FSM agent produced a correct close in 11 of 12 periods over two seeds (mean 0.92, against 0.67): the plain misses are late periods that ended without a report when time ran out, and the FSM's one miss is a final period never started because the run's \$8 budget was exhausted. With 24 turns per period (one seed), the FSM agent ran out of turns in most periods (0.17 vs.\ 0.50).

\paragraph{DuckDB optimizer.}
A Long-Horizon-Terminal-Bench task asks the agent to speed up DuckDB's query optimizer on TPC-H while keeping all 22 queries correct (Claude Sonnet~4.6; ten fresh contexts of at most five turns; the same guidance for both agents). In three of four pairs the plain agent never edited the code, re-reading the guidance in every fresh context. The FSM carried the phase, the agent's own evidence and failed measurements across contexts and reached 1.60--1.64$\times$ speedups in three of four pairs (mean official reward 0.75 vs.\ 0.19), meeting the pre-registered gate on pairs 1--3 ($+0.41$; pair 4 a declared replication); in the remaining pair the FSM agent never changed its failing patch, while the plain agent's minimal edit earned 0.77.

\paragraph{Takeaway.}
In an improvement loop, progress exists only in the record and the measurements: an unmeasured result is not a result, and a context that restarts instead of resuming repeats work. The plain agents failed there (the relabeled ledger, the plan never executed, the periods never reached); the FSM kept the loop's protocol intact: claims wait for measurements, a failed measurement becomes an obligation for the next context, and contexts resume.

\section{Discussion}

\paragraph{When does enforcement help?}
The FSM helped where the plain agent would otherwise abandon part of its instructions, under token pressure, by batching, by handing off a customer, by losing the record of a loop; where the agent largely held the procedure (Qwen3.8-flash once reminded and resumed, Sonnet~4.6 on easy SWE-bench tasks), the gain was small or absent and cost time and tokens. How much a plain agent drifts is therefore a practical test of whether enforcement will pay off.

\paragraph{Close every route, and specify them carefully.}
Because drift moves to whichever route remains open, enforcement has to cover every route; the one our static permissions left open was found in three of eight sessions, and item-bound permissions would close it. Enforcement is also only as good as its specification: over-strict checks cost Retail tasks, and a tight turn budget left the month-end FSM agent no room for its checks. A procedure that matters over a long run, skill or SOP, should therefore come with gates on tool calls, transitions and turn endings, not only text; writing such a workflow is mostly transcription, and the guards are the real work (Appendix~\ref{app:authoring}).

\section{Conclusion}

Agents that receive their procedure once drift from it over long unattended runs, and restated instructions do not stop them. An executable FSM that holds the workflow state outside the context and enforces it at tool calls, transitions and turn endings kept agents on procedure, with gains tracking the plain agent's drift, and extended to debugging, customer service and improvement loops at 1.2 to 3.5 times the plain agent's time or cost where both were measured; because the runtime, not the model, supplies this reliability, small models gain a path to long-horizon procedural work.

\section*{Limitations}

\paragraph{Evaluation Scope.}
Our evaluation demonstrates that executable state enforcement can improve long-horizon procedural fidelity, but several limitations remain. 
The main experiment focuses on a single SOP task, one agent runtime, and a limited set of models. 
Although we observe consistent improvements across seeds and additional domains, broader evaluation over diverse procedures, runtimes, and agent architectures is needed to better characterize the general applicability of the approach.

\paragraph{Decision Quality.}
The FSM in our framework enforces how an agent executes a procedure rather than improving the model's underlying reasoning or decision quality. 
Therefore, stronger procedural compliance does not necessarily translate into better final decisions when the model lacks domain knowledge or makes incorrect judgments within an allowed state. 
Future work could explore combining executable enforcement with mechanisms that improve decision-making capability.

\paragraph{Workflow Specification.}
Executable workflows require explicit specifications and carefully designed guards. 
Poorly specified constraints may introduce unnecessary restrictions or fail to capture important edge cases. 
Developing methods for automatically generating, verifying, and refining workflow specifications remains an important direction.

\paragraph{Execution Overhead.}
Runtime enforcement introduces additional computational overhead due to state tracking and action verification. 
While this overhead enables stronger reliability guarantees for long unattended tasks, future work should investigate more efficient enforcement strategies that preserve procedural guarantees with lower execution cost.

\makeatletter
\ifacl@anonymize\else
\section*{Acknowledgments}
Generative AI tools assisted with literature searching, experiment-summary extraction, paper organization, and drafting and editing this manuscript. The authors remain responsible for checking every numerical claim against the retained experiment artifacts, for reviewing the cited literature, and for the final text and scientific claims.
\fi
\makeatother

\bibliography{custom}

\appendix


\section{Main experiment: protocol details}
\label{app:protocol}

\paragraph{Environment.}
Every session is one \texttt{claude -p} call (Claude Code~2.1.292) with its own empty home directory, the default system prompt and default reasoning settings, built-in tools disabled, and the SOP-Bench tools served by an MCP server: listing the batch and the decisions so far, the five data tools, SOP-Bench's own \texttt{determine\_enforcement\_action}, \texttt{record\_analysis} for the STEP~1--4 record, and \texttt{submit\_decision}. Neither the SOP nor the tools nor the gold labels were modified. The first message is the same template for all conditions, containing the full SOP, the number of accounts and the instruction to process accounts one at a time. The model never sees a budget. External limits, enforced outside the model, were a spending cap per session (\$25 for Claude Haiku~4.5, \$150 for Claude Sonnet~4.6 and a higher cap for Qwen3.8-flash), 6{,}000 turns per call, termination after 20 minutes without a new decision, at most 50 consecutive stop-gate blocks (after which the runtime would let the turn end; this never happened) or 50 resumes per session, termination after 10 resumes without a new decision, and a time limit of 6 hours (12 hours for Sonnet~4.6 and Qwen3.8-flash). No analyzed session reached a spending or time limit; the most expensive Haiku~4.5 session cost \$13.87 and the longest took 115 minutes.

\paragraph{Conditions.}
The reminder of conditions Remind, Remind-loop, FSM, Stop-only and Gates-only consists of the SOP's decision section (5.7) and its list of critical rules (6.1), copied verbatim, and a paragraph stating that the run is unattended, that accounts are processed one at a time, and that there is no need to stop, summarize, pause or rush. It is attached once per account, to the result of the first successful data-tool call for that account. The resume message of the -loop conditions and of Gates-only and the stop-gate message of FSM and Stop-only are produced by the same function: the number of accounts left, the next undecided account, the per-account procedure and the statement that ending the turn saves neither tokens nor context. For FSM only, the stop-gate message ends with one sentence naming the current workflow state, since the other conditions have no state to name. The first message of FSM adds a paragraph that tells the agent to enter the workflow and how to move between states. Stop-only uses a single-state workflow with the stop gate only; Gates-only uses the FSM workflow without the stop gate and is resumed like Remind-loop.

\paragraph{Scoring.}
An account's decision is its first submitted decision. \emph{Investigated}: all five data tools succeeded for the account before its decision. \emph{Recorded}: the last STEP~1--4 record for the account before its decision follows all five retrievals and names the submitted action. \emph{Focused}: no other account's data was retrieved between the first of these retrievals and the decision. Compliant = investigated $\wedge$ recorded $\wedge$ focused ($\mathrm{inv}$, $\mathrm{rec}$ and $\mathrm{foc}$ in Section~\ref{sec:setup}); strict correctness is SOP-Bench's exact label match, and FC counts compliant and strictly correct decisions. Compliance rates are compliant / decided per session, averaged over sessions.

\paragraph{Pre-registration, freeze and deviations.}
The plan fixed conditions, prompts, metric, predictions, statistics and limits; seed~1 was used for development, and the code, workflow and analysis scripts were committed before any formal run, and no run could start if any of these files had changed since. Every later change was recorded before it took effect. The main ones: (i) the diagnostic count of decisions submitted for another account than the last record, added before the formal runs once the permission gap was found; (ii) a correction to the count of stop-gate blocks, which had also counted quotations of the block message inside compaction summaries, applied after the pre-registered analysis had been run and changing no other number; (iii) one session (Stop-only, seed~4) that ended when the model fetched 160 accounts in one turn and overflowed its context, which we classified as a model outcome, not an infrastructure failure, and did not rerun; (iv) for the extension models, written before their runs, the regime rule and predictions (Appendix~\ref{app:predictions}), a check that every response was served by the requested model, and a classification of errors into quota, infrastructure and context overflow.

\paragraph{Infrastructure failures and seeds.}
Models were accessed through a commercial API relay, and every response in every analyzed session was served by the requested model. A session counted as an infrastructure failure when it ended with an API, quota or server error or a crash of the FSM daemon rather than with an action of the model; this was decided from the error returned, never from the session's decisions. Under the pre-registered rule, such a failure voided every session of that seed and a replacement seed was added, numbered from 10 upward; under an amendment written before the seeds it applied to were run, only the interrupted session was rerun from scratch. No voided seed was rerun or reused, including sessions that had completed. The relay account's credit was exhausted twice, which returned a quota error to every running session and voided Qwen3.8-flash seeds 7--12, Claude Sonnet~4.6 seeds 2--4 and 10--12, and the first Haiku~4.5 runs for the comparison with Sonnet (seeds 10--12); the analyzed seeds are therefore 2--9 for Claude Haiku~4.5, 2--6 and 13--15 for Qwen3.8-flash, and 13--15 for Sonnet~4.6. Sonnet sessions were then run one at a time, so the sessions of a seed ran hours apart; two attempts of Sonnet's FSM session on seed~15 were ended by a relay error and by a congestion rule written before it was applied, and rerun from scratch. For the comparison with Sonnet, Haiku~4.5 was run on seeds 13--15 with the same code; these seeds do not enter Table~\ref{tab:sop-main}. No voided or interrupted session enters any analysis.

\section{Pre-registered predictions}
\label{app:predictions}

\begin{table*}[t]
\centering
\small
\renewcommand{\arraystretch}{1.15}
\begin{tabularx}{\linewidth}{@{}p{0.47\linewidth}Xl@{}}
\toprule
\textbf{Prediction} & \textbf{Observed} & \textbf{Verdict}\\
\midrule
\multicolumn{3}{@{}l}{\textit{Claude Haiku~4.5, seeds 2--9}}\\
P1 Prompting is not enough: Plain, Sys, Remind each have fewer than 200 decisions or compliance below 80\% in $\geq$6/8 seeds & 8/8, 8/8, 8/8 & holds\\
P2 Resumption only reroutes: Plain-loop, Sys-loop, Remind-loop each have $\geq$190 decisions and compliance below 80\% in $\geq$6/8 seeds & 8/8, 7/8, 7/8 & holds\\
P3 FSM: FSM decides all 200 with compliance $\geq$95\% in every seed; FSM$-$Remind-loop $\geq$30 on average and positive in $\geq$7/8; conservative contrast $\geq$20 and positive in $\geq$6/8 & contrasts met ($+$90.9, 7/8; $+$70.5, 7/8); compliance 80\% in seeds 4 and 6 & fails\\
P4 No harm: strict accuracy of FSM $\geq$ Remind-loop & 146.5 vs.\ 138.9 & holds\\
P5 Long horizon: FSM$-$Remind-loop in the last window $\geq$ first window & 29.1 vs.\ 2.0 & holds\\
Falsification: mean FSM$-$Remind-loop $\leq$0 or $\geq$4/8 seeds negative & $+$90.9; 1/8 negative & not triggered\\
\midrule
\multicolumn{3}{@{}l}{\textit{Extension models (Plain, Remind-loop, FSM). Regime: Remind-loop compliance $<$80\% drift, $\geq$95\% no drift}}\\
Qwen3.8-flash: no drift (98.1\%), so $|$B$-$Remind-loop$|\leq$10 & $+$12.9, CI [2.6, 28.5] & fails (FSM better)\\
Claude Sonnet~4.6: drift (66.7\%), so FSM$-$Remind-loop $\geq$30 and positive in 3/3 seeds & $+$55.0, positive in 1/3 & fails\\
FSM decides all and is $\geq$95\% compliant & Qwen 99.7\%, Sonnet 100\% & holds\\
Strict accuracy of FSM $\geq$ Remind-loop & Qwen $+$10.5, Sonnet $-$9.3 & holds for Qwen only\\
Falsification (mean $\leq$0 or at least half the seeds negative) & Qwen 1/8, Sonnet 2/3 negative & triggered for Sonnet\\
\midrule
\multicolumn{3}{@{}l}{\textit{Catch-up, seeds 13--15: Haiku~4.5 with FSM vs.\ Sonnet~4.6 ($\geq+$10 exceeds, $\geq-$10 catches up)}}\\
K1: vs.\ plain Sonnet (Plain) & $+$26.0 (seeds: $+$137, $-$19, $-$40) & exceeds\\
K2: vs.\ Sonnet's best baseline & $-$59.0 & does not catch up\\
\bottomrule
\end{tabularx}
\caption{\textbf{Pre-registered predictions and outcomes.}}
\label{tab:predictions}
\end{table*}

Table~\ref{tab:predictions} lists all pre-registered predictions with their outcomes. P1--P5 were written before the main experiment and the extension predictions before any run of the respective model; before running Sonnet~4.6 we also expected it not to drift at all, whereas one Plain and one Remind-loop session batched accounts throughout. Table~\ref{tab:sop-paired} gives the paired comparisons behind the predictions and behind the numbers in Section~\ref{sec:main}: the mean difference per seed, a bootstrap 95\% CI (10{,}000 resamples), the number of seeds with a positive difference, and two-sided sign and Wilcoxon signed-rank tests. The best auto-continue baseline is the highest FC of Plain-loop, Sys-loop and Remind-loop in each seed; the catch-up rows compare Claude Haiku~4.5, rerun on Sonnet's three account orders, with Sonnet's Plain condition and with its best baseline, the higher FC of Plain and Remind-loop in each seed; with three seeds the smallest attainable $p$ is 0.25. P3 failed only on its compliance condition, through the route that the static permissions leave open (Section~\ref{sec:fsm}). Two further observations were not predicted. First, within the first 50 accounts, where every condition still mostly follows the procedure, the FSM's accuracy on fully investigated accounts (77.5\%) was below Remind-loop's (85.4\%), and the FSM agent first remarked on its token budget earlier (median after 13.5 decided accounts, against 26.5). Second, Sonnet~4.6 with the FSM erred mostly by escalating the action (for example, \textit{Temporary Suspension} to \textit{Permanent Account Closure}) while scoring STEP~1 correctly; we have not identified the cause.

\begin{table*}[t]
\centering
\small
\setlength{\tabcolsep}{4pt}
\renewcommand{\arraystretch}{1.15}
\resizebox{\ifdim\width>\textwidth\textwidth\else\width\fi}{!}{%
\begin{tabular}{llccccc}
\toprule
\textbf{Model} & \textbf{Comparison} & \textbf{Mean} & \textbf{95\% CI} & \textbf{Seeds $>0$} & \textbf{Sign $p$} & \textbf{Wilcoxon $p$}\\
\midrule
\multirow{4}{*}{\makecell[l]{Claude\\Haiku~4.5\\(8 seeds)}} & FC: FSM $-$ Remind-loop & $+$90.9 & [46.2, 130.2] & 7/8 & 0.070 & 0.023\\
 & FC: FSM $-$ best -loop baseline & $+$70.5 & [27.4, 114.5] & 7/8 & 0.070 & 0.039\\
 & FC: FSM $-$ Plain & $+$129.9 & [104.1, 151.9] & 8/8 & 0.008 & 0.008\\
 & Correct: FSM $-$ Remind-loop & $+$7.6 & [$-$16.1, 29.8] & 5/8 & 0.727 & 0.484\\
\midrule
\multirow{3}{*}{\makecell[l]{Claude\\Sonnet~4.6\\(3 seeds)}} & FC: FSM $-$ Remind-loop & $+$55.0 & [$-$15.0, 191.0] & 1/3 & 1.000 & 1.000\\
 & FC: FSM $-$ Plain & $+$79.3 & [9.0, 184.0] & 3/3 & 0.250 & 0.250\\
 & Correct: FSM $-$ Remind-loop & $-$9.3 & [$-$15.0, $-$2.0] & 0/3 & 0.250 & 0.250\\
\midrule
\multirow{3}{*}{\makecell[l]{Qwen3.8-\\flash\\(8 seeds)}} & FC: FSM $-$ Remind-loop & $+$12.9 & [2.6, 28.5] & 7/8 & 0.070 & 0.047\\
 & FC: FSM $-$ Plain & $+$28.2 & [2.2, 61.8] & 6/8 & 0.289 & 0.078\\
 & Correct: FSM $-$ Remind-loop & $+$10.5 & [1.2, 23.4] & 7/8 & 0.070 & 0.109\\
\midrule
\multirow{3}{*}{\makecell[l]{Catch-up\\(seeds 13--15)}} & FC: Haiku~4.5 FSM $-$ Sonnet~4.6 Plain & $+$26.0 & [$-$40.0, 137.0] & 1/3 & 1.000 & 1.000\\
 & FC: Haiku~4.5 FSM $-$ Sonnet~4.6 best baseline & $-$59.0 & [$-$75.0, $-$40.0] & 0/3 & 0.250 & 0.250\\
 & Correct: Haiku~4.5 FSM $-$ Sonnet~4.6 Plain & $-$32.7 & [$-$43.0, $-$15.0] & 0/3 & 0.250 & 0.250\\
\bottomrule
\end{tabular}%
}
\caption{\textbf{Paired comparisons over seeds.}}
\label{tab:sop-paired}
\end{table*}

\section{Escape routes}
\label{app:escape}

Table~\ref{tab:sop-escape} counts, for every Haiku~4.5 condition (seeds 2--9; ablations seeds 2--4), how often each route away from the procedure was taken: each cell gives the number of sessions in which the route occurred over the number of sessions, with the mean occurrences per session in parentheses. \emph{Ends turn}: the model ended its turn with accounts left (single-session conditions then stopped; auto-continue conditions were resumed, up to 50 times). \emph{Decides without looking}: no data tool was called for the account before its decision; \emph{partial data}: some but not all five. \emph{Interleaved accounts}: decisions on accounts whose five retrievals were interleaved with another account's. \emph{Submits for another account}: the decision is for a different account than the most recent STEP~1--4 record, the route left open by the FSM's static permissions. The last column is the median number of decided accounts when the model first mentioned tokens, context or budget. Each baseline takes the routes its design leaves open, and FSM, which closes all three, takes only the route its static permissions leave open.

\begin{table*}[t]
\centering
\small
\setlength{\tabcolsep}{4pt}
\renewcommand{\arraystretch}{1.15}
\resizebox{\ifdim\width>\textwidth\textwidth\else\width\fi}{!}{%
\begin{tabular}{lcccccc}
\toprule
\textbf{Condition} & \textbf{\makecell{Ends turn,\\accounts left}} & \textbf{\makecell{Decides\\without looking}} & \textbf{\makecell{Decides on\\partial data}} & \textbf{\makecell{Interleaved\\accounts}} & \textbf{\makecell{Submits for\\another account}} & \textbf{\makecell{First token-\\pressure remark}}\\
\midrule
Plain & 4/8 (0.5) & 4/8 (49.1) & 3/8 (17.5) & 8/8 (65.8) & 5/8 (87.8) & 63\\
Sys & 6/8 (0.8) & 3/8 (32.8) & 1/8 (2.5) & 7/8 (63.6) & 2/8 (32.4) & 66.5\\
Remind & 8/8 (1.0) & 1/8 (20.2) & 0/8 & 6/8 (15.9) & 5/8 (28.2) & 22\\
\midrule
Plain-loop & 6/8 (1.5) & 5/8 (48.6) & 0/8 & 8/8 (122.8) & 4/8 (82.8) & 81\\
Sys-loop & 5/8 (3.0) & 3/8 (38.2) & 2/8 (3.9) & 7/8 (100.0) & 5/8 (88.4) & 69.5\\
Remind-loop & 8/8 (5.5) & 3/8 (29.9) & 5/8 (14.2) & 7/8 (100.2) & 7/8 (105.2) & 26.5\\
\midrule
FSM & 0/8 & 0/8 & 0/8 & 3/8 (10.5) & 3/8 (10.5) & 13.5\\
\midrule
Stop-only & 0/3 & 2/3 (108.3) & 1/3 (1.7) & 3/3 (29.3) & 3/3 (127.3) & 20\\
Gates-only & 3/3 (44.7) & 0/3 & 0/3 & 1/3 (13.7) & 1/3 (13.7) & 13\\
\bottomrule
\end{tabular}%
}
\caption{\textbf{How each condition left the procedure} (Claude Haiku~4.5).}
\label{tab:sop-escape}
\end{table*}

\section{Details of the analyses}
\label{app:analyses}

\subsection{Analysis 1: SWE-bench Verified}
\label{app:analysis-swe}

Table~\ref{tab:analysis-swe} gives resolved tasks, patches that add a regression test, and median wall-clock time and mean cost per task, which were recorded for Easy-100 only; one run per task and agent.

\paragraph{Agents.}
Both agents run Claude Code with the same repository, model, budget and time limit per task. The plain prompt lists four steps (investigate; root cause, with ``do not modify source code yet''; fix; verify) and asks for \texttt{investigation.md} and \texttt{root\_cause.md} to be committed. The FSM agent's prompt invokes a debugging skill whose workflow has states \textsc{Investigate}, \textsc{RootCause}, \textsc{Fix} and \textsc{Exit}; its transitions require a clean working tree, the corresponding markdown file, and no change to Python sources before \textsc{Fix}. The prompt wording thus differs between the two agents, while the procedure is the same. Neither agent is asked to write tests.

\paragraph{Task sets and runs.}
Hard-100 combines two 50-task samples drawn by the same heuristic, which scores task metadata (moderate difficulty, issue and gold-patch length, one to six failing tests, root-cause keywords) and balances tasks across repositories; two tasks were carried over from an earlier pilot in which the FSM had been useful on them, and in Hard-100 one was resolved by both agents and the other by neither, so they contribute nothing to the difference. The selection was not pre-registered. It was run in June 2026 through Claude Code's \texttt{sonnet} alias, which referred to Claude Sonnet~4.6 at the time, with a wall-clock limit per task; no record of that run names the exact model version, and time and cost were not preserved. Easy-100 is a random sample (seed 20260918) of 100 tasks annotated as ``$<$15 min fix'' that had not been used before. It was run in September 2026 with Claude Sonnet~4.6, a \$5 budget and a 1{,}200\,s limit per task; sessions interrupted by relay quota errors were resumed, the retry being decided by the error alone. It was planned as a non-inferiority test with a 5-point margin; with 100 tasks the interval is too wide to establish non-inferiority at that margin, and seven tasks that neither agent resolved may have environment problems, which affects the absolute counts rather than the comparison. We report one run per task and agent, and every patch was scored with the official harness on a clean image.

\paragraph{Process measures.}
``Procedure followed'' requires the investigate, root-cause and fix phases to have been completed in order according to the phase log derived from the commit history; we evaluate it on the 98 plain and 99 FSM Easy-100 attempts that produced a patch. ``New tests'' counts patches that add a function named \texttt{test\_*}; SWE-bench replaces test files with the gold tests when scoring, so this measure does not affect resolution. On Easy-100, the plain and FSM agents had 5 and 5 timeouts, 2 and 1 empty patches, and 37.9 and 48.6 turns per task on average over the tasks that finished within the time limit, over which mean cost is also computed. Table~\ref{tab:analysis-swe-detail} gives, per task set and for the two 50-task halves of Hard-100 (97 of its 100 tasks are annotated ``15\,min--1\,h''), the tasks resolved by each agent, the discordant tasks (resolved only with / only without the FSM) with the exact McNemar $p$, patches adding a test function, empty patches, and median time and mean cost per task, recorded for Easy-100 only.

\begin{table*}[t]
\centering
\small
\setlength{\tabcolsep}{4pt}
\renewcommand{\arraystretch}{1.15}
\resizebox{\ifdim\width>\textwidth\textwidth\else\width\fi}{!}{%
\begin{tabular}{llcccccccc}
\toprule
\textbf{Task set} & \textbf{Model} & \textbf{Plain} & \textbf{FSM} & \textbf{Discordant} & \textbf{$p$} & \textbf{New tests} & \textbf{Empty} & \textbf{Time (s)} & \textbf{Cost (\$)}\\
\midrule
Hard-100 & Claude Sonnet~4.6 & 62 & \textbf{69} & 7 / 0 & 0.016 & 24 / 39 & 0 / 0 & -- & --\\
\quad first 50 &  & 34 & \textbf{37} & 3 / 0 & 0.250 & 13 / 20 & 0 / 0 & -- & --\\
\quad second 50 &  & 28 & \textbf{32} & 4 / 0 & 0.125 & 11 / 19 & 0 / 0 & -- & --\\
\midrule
Easy-100 & Claude Sonnet~4.6 & 86 & \textbf{87} & 5 / 4 & 1.000 & 57 / 64 & 2 / 1 & 310 / 381 & 1.30 / 1.73\\
\bottomrule
\end{tabular}%
}
\caption{\textbf{Analysis 1 in detail} (pairs of values are plain / FSM).}
\label{tab:analysis-swe-detail}
\end{table*}

\subsection{Analysis 2: \texorpdfstring{$\tau^2$}{tau2}-bench}
\label{app:analysis-tau2}

Table~\ref{tab:analysis-tau2} lists per domain the tasks passed by each agent, the discordant tasks (passed only with / only without the FSM) and the exact McNemar $p$, with one run per task and agent; panel (a) is the first frozen evaluation in each domain, and panel (b) later domain-specific FSMs re-evaluated on tasks seen during their development.

\paragraph{Setup.}
Both agents used the unmodified benchmark (v1.0.1), gpt-5.4-mini at medium reasoning effort for the agent and the user simulator, the official policy, tools, step limits and seeds, and the same static checklist. The FSM agent also runs our FSM runtime through a Python adapter. In Telecom its workflow has four states (scope, diagnose, retest, close). The runtime binds the customer's phone line and original symptom, keeps a dependency-ordered ledger of unresolved faults, issues one repair at a time, opens a fresh test after every repair, and denies transfers to a human while faults remain, unless public evidence shows an irreversible condition such as a locked SIM. The workflow contains no task identifiers, hidden state, expected actions or rewards. Because the adapter emits many steps itself (62\% of the FSM agent's turns, against 10\% for the control agent, whose only scripted turns are the benchmark's opening greetings), the result measures the FSM-driven controller as a whole.

\paragraph{Evaluation.}
We report one run per task and agent. Telecom tasks are scored by the benchmark's deterministic action and environment checks and involve no language-model judge. In the other domains, where some tasks carry natural-language assertions, the benchmark's default judge (gpt-4.1) was unavailable to us, and both agents were judged by gpt-5.4-mini with the unchanged rubric. Telecom development used only the training split. The 40-task test split was evaluated in three segments of 12, 12 and 16 tasks, in the order given by a hash of the task identifiers; each later segment was started after the previous one had met its gate and extended only the runner's task list: agent, workflow, rules, runtime, model and scoring were identical in all three segments, and no scored task was rerun. Panel~(a) of Table~\ref{tab:analysis-tau2} contains the first frozen evaluation in each domain. Panel~(b) contains later domain-specific FSMs, developed after earlier results in the same domain were known and evaluated on cohorts that include tasks used during development, so they are descriptive; panel~(b) reports the last revision in each domain, and intermediate revisions evaluated on the same cohorts (two in Airline, one each in Retail and Banking) lie between the two panels. In Banking, the entire gain of the revised FSM lies in the 15 tasks whose workflows it encodes (3/15 vs.\ 15/15); the other 82 tasks tie (8 vs.\ 8). Earlier pilots in Airline and Retail (1--9 tasks each) were mixed, and six later extensions were stopped after a few tasks; none of them enters any estimate. Mean conversation time in Telecom was 230\,s for the control agent and 374\,s for the FSM agent.

\subsection{Analysis 3: Staged protocols}
\label{app:analysis-protocols}

Table~\ref{tab:analysis-protocols} summarizes the three protocols; Table~\ref{tab:analysis-protocols-detail} gives the per-campaign, per-period and per-pair detail.

\paragraph{AutoResearch.}
The task tunes a scikit-learn classifier on the digits dataset from a weak SGD baseline, using Claude Code's default Sonnet model (the version was not recorded). Agents may edit only the training script, preferably just its configuration block. The protocol requires 25 trials over eight stages (baseline 1, scheduler 3, optimizer and learning rate 4, regularization 4, feature augmentation 4, throughput 3, combination 3, confirmation 3), each recorded in a results file and a ledger row. A row is valid if it has the required fields, a decision label in \{keep, discard, crash\}, a known stage, a matching axis, a numeric metric, only the allowed file changed, and a recorded candidate commit. The FSM agent's stage transitions check the ledger before the next stage begins. The FSM run of one campaign timed out at 22/25 rows and was resumed once with a confirm-only prompt; without that campaign the totals are 41/75 vs.\ 75/75 valid rows and 1/3 vs.\ 3/3 complete protocols. In a later version in which both agents were given helper scripts that write rows in the required format, both kept valid ledgers and the completion gap narrowed (FSM 7/10, plain 6/10 over ten campaigns): tooling that makes the compliant action the default does part of the enforcement.

\paragraph{Month-end close.}
The synthetic world is a small trading company with three currencies, two warehouses, about 40 SKUs and injected policy events. Each monthly period is a fresh \texttt{claude -p} session (Claude Sonnet~4.6) sharing only workspace files. The FSM workflow's states follow the six steps of the written close procedure, and transitions are gated by a check script that never sees the answers. A period is scored by the fraction of 11 report fields within tolerance. Each run had an \$8 budget, checked before each period starts, so a run could end slightly above it. The plain agent's misses were periods that ended by time-out or turn cap without a final report; timed-out periods record no cost, so its cost is underestimated. A later, unmatched continuation of the FSM run of seed~1 over further periods is not used anywhere in this paper.

\paragraph{DuckDB optimizer.}
The Long-Horizon-Terminal-Bench DuckDB optimizer task provides the DuckDB~1.1.0 source tree, a benchmark command that rebuilds DuckDB and times the 22 TPC-H queries (with a capped number of rebuilds), a plan profiler and a correctness check; the official verifier gives zero reward if any query result changes. Both agents used Claude Sonnet~4.6 with thinking disabled, ten fresh contexts of at most five turns, \$0.80 per context and \$8 in total, the same prompt, the same public optimization guidance, a deterministic data cache and the unchanged verifier. The FSM workflow has four states (implementation, measurement, verification, terminal). It persists the agent's own evidence, a handoff note and the hash of the current patch across contexts, runs the shared measurement once per new patch and the shared correctness check after the speedup gate, and, when a measurement fails, binds the failure to the patch hash and instructs the next context to change the patch. The pre-registered endpoint was the mean reward difference over three pairs, with confirmation requiring all three pairs valid, at least two FSM wins, a mean difference of at least 0.05 and at least two full successes; the fourth pair was declared before it ran as a no-retry replication. Three caveats apply. The design was developed over many earlier versions, each changed after its results were seen, before the pre-registered version was frozen; within that version, the first pair served as a go/no-go check before the other two were run. The FSM agent stops once it reaches the terminal state, while the plain agent always uses all ten contexts, which affects cost and turn comparisons. The verifier gives about 0.77 to a correct patch without speedup, so mean reward mixes such patches with real successes; restricted to pairs 2--4, the mean difference is again $+0.41$ with two full successes. On other tasks of the benchmark (a data-validation audit, grammar fuzzing, Sokoban and others), earlier FSM variants did not outperform their controls; the largest of these studies, a pre-registered ten-pair confirmation on the data-validation audit task, gave a reward difference of $-0.009$ against a checklist control.

\paragraph{Detail table.}
Table~\ref{tab:analysis-protocols-detail} gives (a) per AutoResearch campaign the valid ledger rows (25 required), whether the protocol was completed and the best accuracy; (b) per month-end period the score, which was 1 or 0 in every scored period, zeros being periods that ended without a report by wall-clock timeout or turn cap, and -- a period not started because the \$8 budget was exhausted (in the 24-turn condition the time limit per period was 240\,s instead of 300\,s); and (c) per DuckDB pair the official reward and the verified speedup from the unchanged verifier (all submissions passed the 22-query correctness check), the contexts used and the cost, with pair 4 the declared replication.

\begin{table*}[t]
\centering
\small
\setlength{\tabcolsep}{4pt}
\renewcommand{\arraystretch}{1.15}
\begin{tabular*}{\textwidth}{@{\extracolsep{\fill}}lcccccc@{}}
\toprule
 & \multicolumn{2}{c}{\textbf{Valid ledger rows}} & \multicolumn{2}{c}{\textbf{Complete protocol}} & \multicolumn{2}{c}{\textbf{Best accuracy (\%)}}\\
\cmidrule(lr){2-3}\cmidrule(lr){4-5}\cmidrule(lr){6-7}
\textbf{(a) AutoResearch} & \textbf{Plain} & \textbf{FSM} & \textbf{Plain} & \textbf{FSM} & \textbf{Plain} & \textbf{FSM}\\
\midrule
Campaign 1 & 7/25 & 25/25 & no & yes & 99.33 & 99.33\\
Campaign 2 & 8/25 & 25/25 & no & yes & 99.33 & 99.33\\
Campaign 3 & 8/25 & 25/25 & no & yes & 99.33 & 99.33\\
Campaign 4 & 25/25 & 25/25 & yes & yes & 99.33 & 99.33\\
\bottomrule
\end{tabular*}

\vspace{6pt}
\begin{tabular*}{\textwidth}{@{\extracolsep{\fill}}llcccc@{}}
\toprule
\textbf{(b) Turn cap} & \textbf{Seed} & \textbf{Plain, periods 1--6} & \textbf{FSM, periods 1--6} & \textbf{Mean (plain / FSM)} & \textbf{Cost \$ (plain / FSM)}\\
\midrule
40 & 1 & 1 1 1 0 0 0 & 1 1 1 1 1 -- & 0.50 / 0.83 & 1.81 / 8.64\\
40 & 2 & 1 1 1 1 1 0 & 1 1 1 1 1 1 & 0.83 / 1.00 & 3.82 / 5.61\\
24 & 1 & 1 1 1 0 0 0 & 1 0 0 0 0 0 & 0.50 / 0.17 & 2.27 / 3.54\\
\bottomrule
\end{tabular*}

\vspace{6pt}
\begin{tabular*}{\textwidth}{@{\extracolsep{\fill}}lcccccc@{}}
\toprule
 & \multicolumn{2}{c}{\textbf{Official reward}} & \multicolumn{2}{c}{\textbf{Verified speedup}} & & \\
\cmidrule(lr){2-3}\cmidrule(lr){4-5}
\textbf{(c) DuckDB pair} & \textbf{Plain} & \textbf{FSM} & \textbf{Plain} & \textbf{FSM} & \textbf{Contexts (plain / FSM)} & \textbf{Cost \$ (plain / FSM)}\\
\midrule
1 & 0.00 & 1.00 & -- & 1.64$\times$ & 10 / 4 & 1.25 / 0.78\\
2 & 0.00 & 1.00 & -- & 1.60$\times$ & 10 / 6 & 0.94 / 0.99\\
3 & 0.77 & 0.00 & 1.00$\times$ & -- & 10 / 10 & 1.19 / 1.56\\
4$^{r}$ & 0.00 & 1.00 & -- & 1.63$\times$ & 10 / 8 & 1.07 / 1.33\\
\textit{mean} & 0.19 & 0.75 &  &  &  & 1.11 / 1.17\\
\bottomrule
\end{tabular*}
\caption{\textbf{Analysis 3 in detail.}}
\label{tab:analysis-protocols-detail}
\end{table*}

\section{Writing the workflows}
\label{app:authoring}

Table~\ref{tab:authoring} gives, for the six workflows of this paper as run (for AutoResearch, the version behind the four-campaign result), the number of states and the line counts of the YAML file and of the check code its guards call.

\begin{table*}[t]
\centering
\small
\setlength{\tabcolsep}{4pt}
\renewcommand{\arraystretch}{1.15}
\begin{tabularx}{\textwidth}{@{}llccX@{}}
\toprule
\textbf{Experiment} & \textbf{Workflow} & \textbf{States} & \textbf{YAML lines} & \textbf{Guards and other check code (lines)}\\
\midrule
Main experiment & ReferralAbusePerAccount & 3 & 61 & one guard script with three modes (156); stop-gate message wrapper (95); the YAML and the reminder text are generated from the SOP by a build script (119)\\
SWE-bench Verified & SWEDebug & 4 & 78 & seven inline shell checks (clean tree, notes file exists, no \texttt{.py} change before the fix state); no script\\
$\tau^2$-bench Telecom & TauTelecomTroubleshootingV1 & 4 & 73 & no hard checks, deny rules only; Python adapter: rules (535), runtime bridge (374), controller (377)\\
AutoResearch & AutoResearchSklearn & 10 & 160 & one stage-check script, nine calls (238)\\
Month-end close & MonthlyClose & 7 & 108 & one stage-check script, six calls (177)\\
DuckDB optimizer & DuckDBLongHorizonCausalV21 & 5 & 126 & stage guard (341) and coherence check (62); context-boundary hook (1{,}214) and phase-action runner (262)\\
\bottomrule
\end{tabularx}
\caption{\textbf{Size of the six workflow specifications and their check code.}}
\label{tab:authoring}
\end{table*}

Each workflow mirrors the written procedure state by state (the per-account loop of the SOP, the four debugging steps, the six close steps, the eight tuning stages), its instructions repeat what the procedure says, and its permissions name the tools of that step; writing the YAML was therefore mostly transcription, 61--160 lines of which most is instruction text, and the main experiment's file is generated from the SOP. The guards are the work, and the useful ones check the \emph{process}, never the answer: tool names and call order in the main experiment, \texttt{git} state and the existence of notes in debugging, required fields and arithmetic consistency of the agent's own ledgers in the close and tuning tasks. Two workflows mark the other end of the range: the Telecom YAML has no hard checks, and the fault ledger and goal binding live in a Python adapter and controller of about 1{,}300 lines, which is why Section~\ref{sec:analysis-tau2} credits the FSM-driven controller as a whole; in DuckDB the guards are modest but the context-boundary hook that carries evidence, the patch hash and failed measurements across fresh contexts is 1{,}214 lines, the compound part of that treatment. The failures we hit were specification errors, not runtime errors: over-strict guards (the first Retail workflow rejected order changes that policy allowed and lost 15 against 9 tasks), a turn budget too tight for the checks (24 turns per period in the month-end close), and a permission model that cannot refer to the current item (the open route of the main experiment). The repositories record the iterations: one Telecom, 13 Airline, 17 Retail and 22 Banking workflow files in the $\tau^2$ adapter, 28 earlier versions of the DuckDB experiment (12 pilots and 16 versions of the design that was finally pre-registered), and three workflow versions of the main experiment before the pre-registered one; authoring time was not recorded. When a workflow was wrong, it was wrong in a guard that encoded an assumption about the task instead of a property of the procedure, and the fix was to check less, not more.

\section{Reproducibility}
\label{sec:reproducibility}

For the main experiment we will release the harness, the workflow specifications, guard and hook scripts, the reminder and resume texts, the scoring and analysis code, the pre-registration records, the change log and all session logs. Every number and table in Section~\ref{sec:main} and Appendices~\ref{app:protocol}--\ref{app:escape} is generated from the session logs by these scripts. For the analyses we will release the prompts, workflow specifications, adapters, run summaries and evaluation results.

\end{document}